\documentclass[letterpaper, 10pt, conference]{ieeeconf}

\IEEEoverridecommandlockouts

\usepackage{amsmath,amssymb,amsfonts}
\usepackage{algorithmic}
\usepackage{graphicx}
\usepackage{textcomp}
\usepackage{xcolor}
\usepackage{multirow}
\usepackage{comment}
\usepackage{colortbl}    
\usepackage{color,soul}
\usepackage{threeparttable}
\usepackage{booktabs}
\usepackage{amssymb}
\let\labelindent\relax
\usepackage{enumitem}
\usepackage{url}
\usepackage{array}    
\usepackage{tablefootnote} 
\usepackage{bm}
\usepackage{cite}
\usepackage{multirow}
\usepackage{float}
\usepackage{tabularx}
\usepackage{graphicx}
\usepackage{hyperref}

\makeatletter

\newcommand{\Rmnum}[1]{\expandafter\@slowromancap\romannumeral #1@}

\makeatother

\begin{document}
\title{\LARGE \bf 
{MAGE: A Multi-task Architecture for Gaze Estimation \\ with an Efficient Calibration Module}}

\author{Haoming Huang, Musen Zhang, Jianxin Yang, Zhen Li, Jinkai Li, Yao Guo, \emph{Member}, \emph{IEEE} 
\thanks{This work was supported by National Natural and Science Foundation of China under grant 62203296, the Shanghai Jiao Tong University Medicine-Engineering Interdisciplinary Center Project, the Shanghai Pilot Program for Basic Research - Shanghai Jiao Tong University (No. 21TQ1400203), and the Science and Technology Commission of Shanghai Municipality under Grant 20DZ2220400. 
(\textit{Corresponding authors: Jinkai Li, Yao Guo})}
\thanks{H. Huang, M. Zhang, J. Yang, J. Li, Y. Guo are with the Institute of Medical Robotics, School of Biomedical Engineering, Shanghai Jiao Tong University, Shanghai, China. (\texttt{\{huanghaoming2004, musen\_zhang, jianxinyang, lijinkai, yao.guo\}@sjtu.edu.cn}).} 
\thanks{Z. Li is with Chinese University of Hong Kong (Shenzhen), China. (\texttt{lizhen@cuhk.edu.cn})} }

\maketitle
\thispagestyle{empty}
\pagestyle{empty}

\begin{abstract}
Eye gaze can provide rich information on human psychological activities, and has garnered significant attention in the field of Human-Robot Interaction (HRI). 
However, existing gaze estimation methods merely predict either the gaze direction or the Point-of-Gaze (PoG) on the screen, failing to provide sufficient information for a comprehensive six Degree-of-Freedom (DoF) gaze analysis in 3D space. Moreover, the variations of eye shape and structure among individuals also impede the generalization capability of these methods.
In this study, we propose MAGE, a Multi-task Architecture for Gaze Estimation with an efficient calibration module, to predict the 6-DoF gaze information that is applicable for the real-word HRI.
Our basic model encodes both the directional and positional features from facial images, and predicts gaze results with dedicated information flow and multiple decoders.
To reduce the impact of individual variations, we propose a novel calibration module, namely Easy-Calibration, to fine-tune the basic model with subject-specific data, which is efficient to implement without the need of a screen.
Experimental results demonstrate that our method achieves state-of-the-art  performance on the public MPIIFaceGaze, EYEDIAP, and our built IMRGaze datasets.

\end{abstract}

\begin{figure*}[t]
    \centering
    \includegraphics[width=0.99\linewidth]{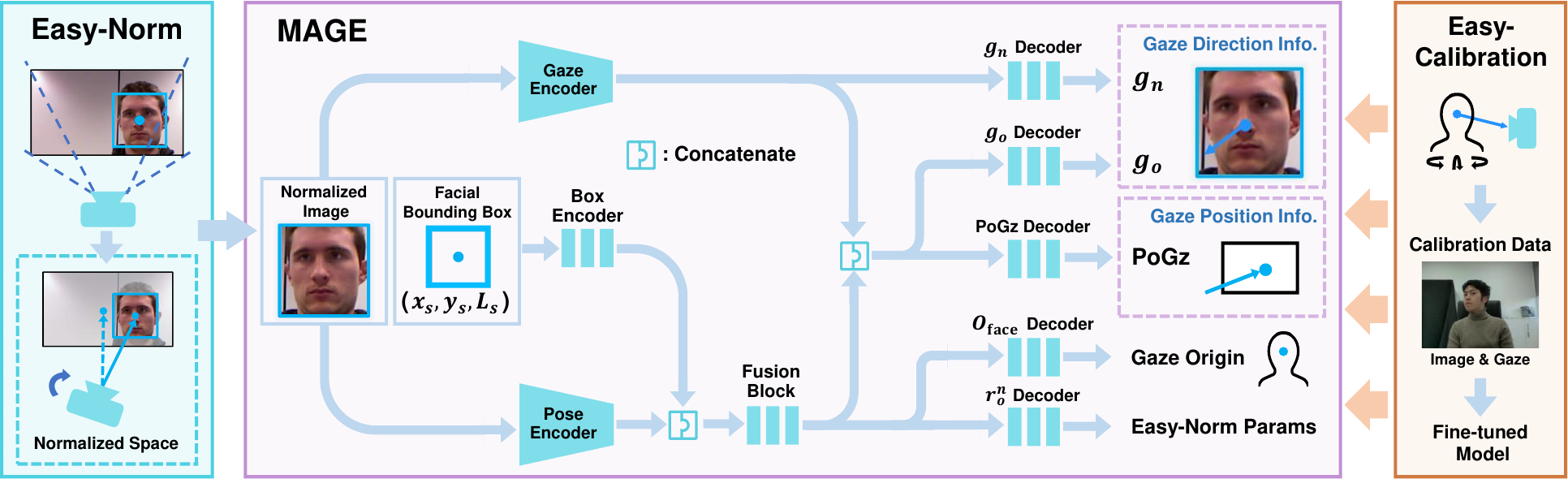} 
    \vspace{-4pt}
    \caption{The overview of MAGE, our proposed multi-task network architecture for human gaze estimation. The model takes the normalized RGB image with facial bounding box provided by the Easy-Norm module as input, and predicts the 6-DoF gaze with complete directional and positional information. The Easy-Calibration module efficiently supplies calibration data for fine-tuning MAGE.}
    \label{fig:the whole pipeline} 
    \vspace{-10pt}
\end{figure*}

\section{Introduction}
Recent advancements of intelligent robotics have expanded their applications from industrial settings to critical healthcare sectors \cite{guo2022medical,aymerich2023socially}. Social robots, in particular, have gained increasing attention for their role in diagnosis, early intervention, and rehabilitation of individuals with neurological disorders. 
These applications heavily rely on robust yet accurate Human-Robot Interaction (HRI) capable of interpreting human behavioral cues to aid decision-making \cite{guo2021human,zhang2022engagement}. Among various HRI interfaces, eye gaze stands out as a pivotal modality. The gaze direction and movement patterns convey rich information about attention, cognitive states and neurological conditions~\cite{guo2021eye}. Therefore, the estimation of human gaze is crucial for robots to effectively understand and respond to their intentions, where most recent works focus on developing vision-based gaze estimation methods that can be easily applied in real-world scenarios.

Currently, non-contact gaze estimation primarily relies on analyzing facial and eye images captured from external viewpoints. Based on the output format of gaze estimation, existing methods can be divided into two categories \cite{cheng2024appearance}: gaze direction prediction and Point-of-Gaze (PoG) prediction.

\textbf{1) Gaze direction prediction}. These methods generate a 3D vector originating from the ocular center to the fixation target, indicating the gaze direction. Existing approaches \cite{li2015gaze,zhang2017mpiigaze, Cheng_2018_ECCV} typically take facial or eye images as input, employ deep neural networks to extract gaze-related features and directly predict the 3D gaze vector. For continuous gaze tracking, some studies \cite{BM:palmero2018recurrent, temporal_cvprw24} leveraged temporal module to exploit continuity constraints for gaze direction estimation.
However, these methods are facing two major challenges: 
First, data-driven deep learning approaches impose strict requirements both on data quantity and quality, thus necessitating data normalization to mitigate confounding factors such as head pose and illumination variation. This requires precise head pose and depth information, limiting use in RGB-only camera systems. Second, while a 3D line requires six Degrees of Freedom (6-DoF) for complete positional and directional information, these methods only estimate the directional component of gaze, resulting in incomplete representations and inadequate performance in HRI.    

\textbf{2) PoG prediction}.
These approaches use deep neural networks to determine the 2D coordinates of PoG on a fixation target plane (typically a screen).\cite{Krafka_2016_CVPR} introduced a large-scale 2D gaze dataset GazeCapture for model training, and proposed the use of face grid to encode head location information. \cite{BM:bao2021adaptive} used an adaptive feature fusion network that integrates features from both facial and eye images for PoG estimation, and achieved state-of-the-art (SOTA) performance on MPIIFaceGaze dataset \cite{Zhang_2017}. However, these methods provide only 2D positional information of gaze.
Furthermore, parameter variations across different devices necessitate the retraining of models, limiting their applicability in natural scenes.

In addition, individual differences in eye anatomy, particularly discrepancies of kappa angle (i.e., the angle between the visual axis and the optical axis) introduce significant challenges like high variance, instability, and domain shifts in gaze estimation. While calibrations are used in existing studies \cite{zhang2018training, calibration_icra17, differential} to address this challenge, the dependence of the interaction screen and the need for manual annotations further impede their real-world applicability, and can compromise the estimation accuracy.

To address the aforementioned challenges, we propose MAGE, a Multi-task Architecture for Gaze Estimation with an efficient calibration module. Our method takes an RGB image with facial bounding box as input, to predict the 6-DoF gaze in the Camera Coordinate System (CCS).
Firstly, we introduce a simplified data normalization method for gaze estimation task, namely Easy-Norm, which only requires the facial bounding box information. Our basic model utilizes the normalized facial image to produce multi-task outputs, including the 3D gaze vector and the gaze point on the $XY$-plane of CCS (denoted as PoGz), which respectively describe the directional and positional information of human gaze. 
To enhance the gaze estimation performance on different subjects, we propose a novel calibration module, namely Easy-Calibration, which is efficient to apply without the need of a screen. The calibration data are utilized to fine-tune our basic model and mitigate the domain shift problem.
Experimental results demonstrate the superior performance of our method against the SOTA gaze estimation methods.
To sum up, the contributions of this paper are three folds:
\begin{enumerate}
\item We propose a novel multi-task architecture, i.e., MAGE, for human gaze estimation, 
which predicts the complete 6-DoF gaze with directional and positional attributes.

\item We propose an efficient calibration module that is easy to implement without the need of a screen. This module fine-tunes MAGE with the calibration data, hence adapting the model to the individual variations.

\item Experimental results demonstrate the SOTA performance of our method on both the gaze direction prediction and the PoG prediction tasks.
\end{enumerate}

\section{Methodology}

In this section, we introduce our proposed gaze estimation method, MAGE, which is composed of a multi-task network architecture, a simplified data normalization module named Easy-Norm, and an efficient calibration module named Easy-Calibration,
as illustrated in Fig. \ref{fig:the whole pipeline}. Our method utilizes
the RGB image with facial bounding box
as input
to simultaneously predict the gaze vector in the CCS and 
the point-of-gaze on the $XY$-plane of CCS, denoted as PoGz. 
Additionally, our method also outputs 
the position of gaze origin (defined as face center in this study) and the parameters for gaze data normalization.

\subsection{Gaze Data Normalization}

\begin{figure}[t]
    \vspace{+4pt}
    \centering
    \includegraphics[width=0.98\linewidth]{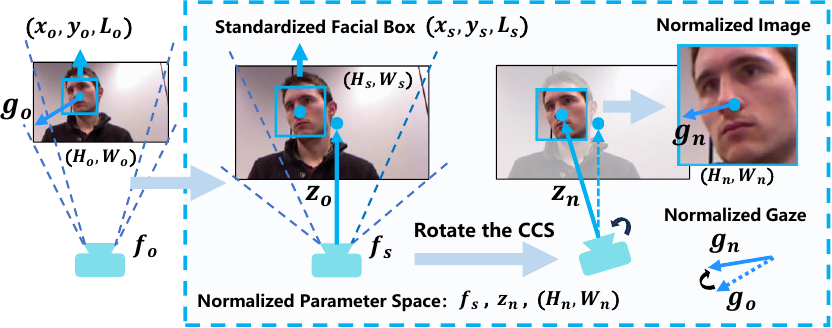}
    \vspace{-4pt}
    \caption{The pipeline of Easy-Norm, involving two steps: (1) Standardize the camera parameters, and transform the image and facial bounding box accordingly. (2) Aligning the oCCS by rotating the z-axis toward the face center ; $\bm{g}_n$ is derived by inversely rotating the $\bm{g}_o$ accordingly. }
    \label{fig:normalization}
\end{figure}

As described in \cite{zhang2018revisiting}, the most commonly used method of gaze data normalization requires the accurate head pose and depth information, making it difficult to apply in the scenarios with only an RGB camera. In this study, we introduce the Easy-Norm, a simplified normalization method using only the facial bounding box on the image, as illustrated in Fig. \ref{fig:normalization}. 
The Easy-Norm involves the following 2 steps:

\textbf{Step 1.} Standardize the camera intrinsic parameters into focus length $f_s$, with the principal point on the image center, and resize the image into ${H_s \times W_s}$.
The facial bounding box $(x_o,y_o,L_o)$ is converted into $(x_s,y_s,L_s)$. It should be noted that, in this step, the original CCS (denoted as oCCS) is not changed, as well as the original gaze vector $\bm{g}_o$.

\textbf{Step 2.} 
The facial region within the bounding box is cropped and converted into a fixed size ${H_n \times W_n}$ for model input. 
The optical axis $\bm{z}_o$ of the camera is aligned from the principal point  $(x_p, y_p)$
to the face center $(x_s,y_s)$, resulting in the normalized CCS (denoted as nCCS). The rotation vector $\bm{r}_o^n = \theta \cdot \bm{r}$ for transforming the oCCS into nCCS can be derived by:
\vspace{-4pt}
\begin{equation}\label{rotation vector}
\begin{cases}
    \bm{r} = \bm{z}_o \times \bm{z}_n \\
    \theta = \arccos(\bm{z}_o \cdot \bm{z}_n) \\
    \bm{z}_n= \frac{\bm{z}_s}{\|\bm{z}_s\|} \\
    \bm{z}_s=[x_s-x_p,y_s-y_p,f_s]^\mathrm{T} \\
    \bm{z}_o=[0,0,1]^\mathrm{T}
\end{cases}
\text{,}
\end{equation}
where $\bm{z}_n$ and $\bm{z}_o$ denote the $Z$-axis of the nCCS and oCCS, respectively. The normalized gaze vector $\bm{g}_n$ is:
\begin{equation}
    \bm{g}_n = \bm{R}_o^n \cdot \bm{g}_o
    \label{g_n}
\text{ ,}
\end{equation}
where $\bm{R}_o^n$ is the rotation matrix derived from $\bm{r}_o^n$.



\subsection{Model Architecture}\label{sec:model architecture}
The architecture of our model, MAGE, is illustrated in Figure \ref{fig:the whole pipeline}. The main idea of the structure is to extract the comprehensive gaze information, including both position and direction, from the normalized image along with the bounding box given by the Easy-Norm module. 

Consequently, the encoding phase of our model can be viewed as comprising two main branches. The first branch involves the normalized image being fed into the Gaze Encoder, which is specifically designed to extract features related to the eye region, thereby providing the directional gaze information. The second branch consists of the normalized image being input into the Pose Encoder and the bounding box being input into the Box Encoder. The Pose Encoder is responsible for extracting implicit spatial information regarding the head pose, while the Box Encoder directly supplies information related to the position of face center and the rotation of coordinate system during the normalization process. The Fusion Block combines these two parts and provides positional information regarding the gaze origin.

In the decoding phase of the model, the directional information from the Gaze Encoder is directly decoded to yield the normalized gaze $\bm{g_n}$. Apart from the directional information, the prediction of the original gaze $\bm{g_o}$ requires the information related to the rotation of coordinate system, and the PoGz (described in Section \ref{sec:pogz}) necessitates positional information. Hence, their decoders concatenate the features from both input branches to produce the final outputs. The above decoders provide the complete gaze information we require. Additionally, the model outputs the coordinates of face center and the parameters used for data normalization, explicitly constraining the model to achieve accurate positional information extraction. In our model, the Gaze and Pose Encoders utilize ResNet-18\cite{he2015deepresiduallearningimage}, while the remaining components consist of a combination of multiple linear layers.




\subsection{Definition of PoGz}
\label{sec:pogz}
\begin{figure}[t]
    \centering
    \includegraphics[width=0.83\linewidth]{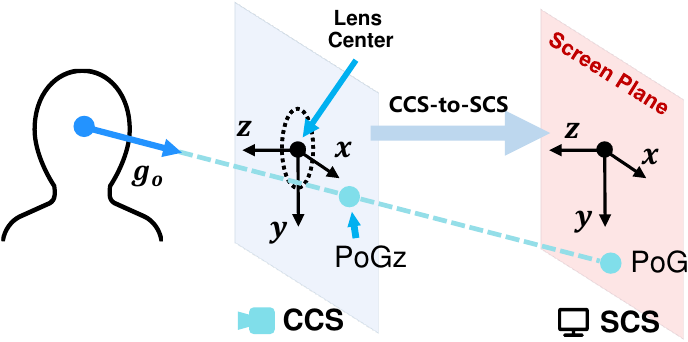}
    \caption{The definition of PoGz is the intersection point of the gaze vector $\bm{g}_o$ and the $XY$-plane of the CCS. Then, the coordinates of PoG in the SCS can be derived from PoGz.} 
    \label{fig:pogz} 
\end{figure}

As show in Fig. \ref{fig:pogz}, we define the PoGz as  
the intersection point of the original gaze vector $\bm{g}_o$ and the $XY$-plane of oCCS. The PoGz is jointly determined by both the gaze direction and the position of gaze origin. Therefore, the prediction of PoGz combines information from the normalized image and the facial box bounding, 
which offers the following two main benefits.
On one hand, the PoGz provides the positional information of human gaze in the 3D space, which is absent in the common task of gaze direction estimation.
On the other hand, compared with the traditional PoG, our defined PoGz is a device-independent representation of gaze, which can enable unified model training on different datasets, and enhance the scalability and generalization capability of the model.
In addition, PoG and PoGz can be converted into each other. 
Let $\beta$ be the $XY$-plane of the screen coordinate system (SCS), $\bm{R}_{c}^{s}$ and $\bm{t}_{c}^{s}$ be the known rotation matrix and translation vector between the oCCS and the SCS, we can derive PoG by:
\begin{equation}\label{pogz2pog}
\begin{cases}
     O_{\mathrm{PoGz}}^s = \bm{R}_{c}^{s} \cdot O_{\mathrm{PoGz}} + \bm{t}_{c}^{s} \\
     \bm{g}^s = \bm{R}_{c}^{s} \cdot \bm{g}_o + \bm{t}_{c}^{s} \\
     l^s = O_{\mathrm{PoGz}}^s + \lambda \cdot \bm{g}^s \text{, } \mathrm{where} \ \lambda \in \mathbb{R} \\
     \mathbf{PoG} = l^s \cap \beta\;\\
\end{cases}
\text{,}
\end{equation}
where $O_{\mathrm{PoGz}}$ and $O_{\mathrm{PoGz}}^s$ are the coordinates of PoGz in the oCCS and SCS, respectively. The $\bm{g}^s$ is the gaze vector in the SCS, and $l^s$ refers to the gaze line in the SCS represented by parametric equation. 


\subsection{Multi-task Output}
\subsubsection{Prediction of Gaze}
As mentioned in Section \ref{sec:model architecture}, our model outputs gaze information including $\bm{g}_n$, $\bm{g}_o$ and PoGz.
For $\bm{g}_n$ and $\bm{g}_o$ prediction, the vectors are normalized to unit vectors and then projected into 2D Euler angle representation formulated as $\bm{g}_\mathrm{2d} = (\phi_\mathrm{yaw}, \theta_\mathrm{pitch})$, where $\phi_\mathrm{yaw}$ and $\theta_\mathrm{pitch}$ denote the horizontal and vertical gaze angles, respectively. These tasks are supervised by the mean absolute error (MAE) loss function:
\begin{equation}
\mathcal{L}_{\bm{g}} = \|\bm{g}_\mathrm{2d} - \hat{\bm{g}}_\mathrm{2d}\|_1
\text{ ,}
\end{equation}
where $\hat{\bm{g}}_\mathrm{2d}$ is prediction and $\bm{g}_\mathrm{2d}$ is ground truth.
The PoGz estimation task is supervised by:
\begin{equation}
\mathcal{L}_{\mathrm{PoGz}} = \|O_{\mathrm{PoGz}} - \hat {O}_{\mathrm{PoGz}}\|_1
\text{ .}
\end{equation}
Since PoGz is a point on the $XY$-plane of the oCCS, we drop the constant zero z-component in practical application. The same treatment is applied to PoG.

\subsubsection{Prediction of Normalization Parameter and Gaze Origin} 
In addition to the gaze part we need, the model also outputs the rotation vector 
$\bm{r}_o^n$ for Easy-Norm and the gaze origin $O_{\mathrm{face}}$, both requiring the positional information. For the gaze origin, our model first predicts its depth value, and then calculate the complete coordinates with the face center on the image. We employ the following constraints to assist the model in better encoding the positional information:
\begin{equation}
\begin{cases}
     \mathcal{L}_{\bm{r}_o^n} = \|\bm{r}_o^n -  \hat{\bm{r}}_o^n\|_1 \\
     \mathcal{L}_{\mathrm{face}} = \| {O}_{\mathrm{face}} -\hat {O}_{\mathrm{face}}\|_1
\end{cases}
\text{.}
\end{equation}
Since the self-rotation component of oCCS around the optical axis $\bm{z}_o$ is always zero in the Easy-Norm, we drop the z-component thus $\bm{r}^n_o$ becomes a 2D output.
\subsubsection{Multi-task Constraints\label{sec:multi}}
With the outputs described above, we construct the overall constraints for the model to predict the complete gaze information more precisely:
\begin{equation}
\mathcal{L} = \lambda_{\bm{g}_n}\mathcal{L}_{\bm{g}_n}+  \lambda_{\bm{g}_o}\mathcal{L}_{\bm{g}_o}
+ \lambda_{\mathrm{PoGz}}\mathcal{L}_{\mathrm{PoGz}}
+  \lambda_{\bm{r}_o^n}\mathcal{L}_{\bm{r}_o^n}+  \lambda_{\mathrm{face}}\mathcal{L}_{\mathrm{face}}
\text{ ,}
\end{equation}
where $\lambda_{\bm{g}_n}$, $\lambda_{\bm{g}_o}$, $\lambda_{\mathrm{PoGz}}$, $\lambda_{\bm{r}_o^n}$ and $\lambda_{\mathrm{face}}$ are the weights of different constraints.



\begin{figure}[t]
    \centering
    \includegraphics[width=0.98\linewidth]{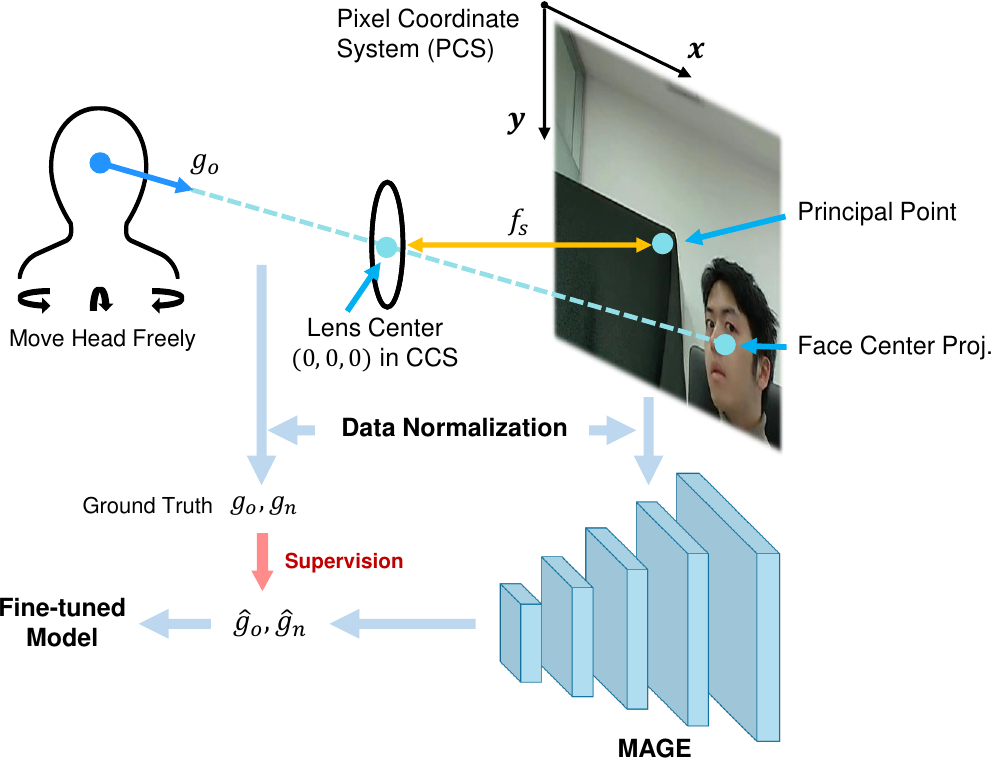}
    \vspace{-5pt}
    \caption{The pipeline of the Easy-Calibration module. Subjects gaze at the camera lens center and move their head to capture multi-angle images. The ground truth of gaze vector $\bm{g}_o$ on these images can be derived as the unit vector pointing from the face center projection to the lens center in the CCS, which is then utilized to fine-tune the basic model.}
    \label{fig:calibration} 
    \vspace{+2pt}
\end{figure}

\subsection{Easy-Calibration Module}

Existing calibration methods for human gaze estimation are typically screen-based ones \cite{zhang2018training, calibration_icra17}, which are not applicable in real-world scenarios, e.g., interacting with some physical objects. In this work, we propose a novel calibration module, namely Easy-Calibration, by extending our method in \cite{li2023easygaze3d} to mitigate the domain shift problem and enhance the gaze estimation performance of the basic model.

As shown in Fig. \ref{fig:calibration}, the collection of the calibration data requires the subjects gaze at the camera lens center, i.e., the origin of the CCS, and move their head freely, which is efficient to apply. At the same time, the frames captured by the camera are collected as the calibration images of each subject. The ground-truth of gaze vector $\bm{g}_o$ on these images can be derived by:
\begin{equation}
\begin{cases}
    \bm{g}_o = -[x_{s}^c,y_{s}^c,f_s]^\mathrm{T}  \\
    x_{s}^c=  (x_s-x_p) \cdot k \\
    y_{s}^c = (y_s-y_p) \cdot k
\end{cases}
\label{eq:cali-gt}
\text{,}
\end{equation}
where $(x_s,y_s)$ and $(x_p,y_p)$ denote the face center projection and the principal point on the standardized image, $k$ is the pixel-to-millimeter conversion factor, and $f_s$ is the focus length of the camera. The $(x_{s}^c,y_{s}^c,f_s)$ is the coordinate of face center projection in the CCS, which is on the gaze line when the subject gazes at the lens center, as shown in Fig. \ref{fig:calibration}. Then, the ground-truth of gaze vector $\bm{g}_o$ can be derived as Eq. \ref{eq:cali-gt}.
The calibration data are utilized to fine-tune our basic model for each subject, which can improve the accuracy of gaze estimation and adapt the model to the individual variations.


\section{Experiments}

\subsection{Dataset}

\begin{figure}[t]
    \centering
    \includegraphics[width=\linewidth]{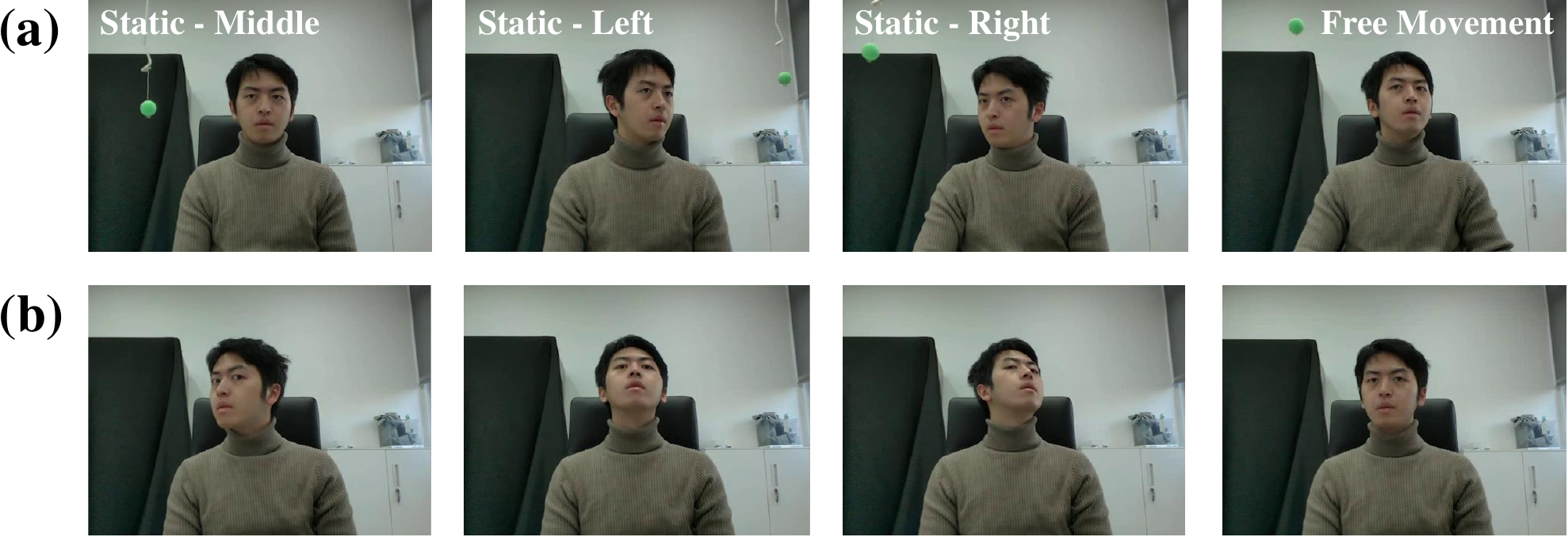}
    \vspace{-17pt}
    \caption{Illustration of IMRGaze dataset. (a) General data in IMRGaze are collected under three static head poses (middle, left, and right) and free movement. (b) Calibration data in IMRGaze, in which the subjects gaze at camera lens center while moving the head freely.}
    \label{fig:imr} 
    \vspace{+6pt}
\end{figure}

\begin{table}[t]
    \centering
    \caption{Distribution of gaze direction and head pose in IMRGaze general data.}
    \renewcommand\arraystretch{1.3}
    \vspace{-5pt}
    \begin{tabular}{lccc}
    \hline\hline
    Attribute (rad)         & Range & Mean & Std \\  \hline
    Gaze Yaw &  [-0.86, 0.97] & 0.01 & 0.27     \\   
    Gaze Pitch &  [-0.69, 0.84] & -0.01 & 0.23  \\ \hline
    Head Pose Yaw &  [-1.45, 1.12] & 0.03 & 0.29     \\   
    Head Pose Pitch &  [-0.77, 0.72] & -0.01 & 0.12  \\
    \hline
    \end{tabular}
    \label{tab:distribution}
    \vspace{+2pt}
\end{table}

\begin{table*}[t]
    \centering
    \caption{Comparison of gaze direction and PoG estimation on three publicly available datasets}
    \label{tab:BMscreen}
    \renewcommand\arraystretch{1.2}
    \begin{tabular}{lccccccc}
    \hline\hline
        \multirow{2}{*}{Model}                 & \multirow{2}{*}{Gaze Dir. Pred.} & \multirow{2}{*}{PoG Pred.} & \multicolumn{2}{c}{MPIIFaceGaze} & \multicolumn{2}{c}{EYEDIAP-Screen} & EYEDIAP-Float  \\  
                                               & & & Gaze Dir. (\textdegree) & PoG (mm) & Gaze Dir. (\textdegree) & PoG (mm) & Gaze Dir. (\textdegree) \\ \hline
        iTracker \cite{Krafka_2016_CVPR}       & & \checkmark & N/A & 76.70 & N/A & 101.30 & N/A  \\ 
        AFF-Net \cite{BM:bao2021adaptive}      & & \checkmark & N/A & 42.10 & N/A & 92.50 & N/A  \\     
        GazeTR \cite{BM:cheng2022gaze}         & \checkmark & & 4.00 & N/A & 5.17 & N/A & 6.23  \\ 
        AGE-Net \cite{BM:shi2024agent}         & \checkmark & & 3.82 & N/A & 4.86 & N/A & 5.57  \\
        RCNN \cite{BM:palmero2018recurrent}    & \checkmark & & 4.10 & N/A & 5.31 & N/A & 5.19  \\ 
        L2CS-Net \cite{abdelrahman2023l2cs}       & \checkmark & & 3.92 & N/A & 5.57 & N/A & 6.68  \\ 
        \textbf{MAGE (ours)}                   & \checkmark & \checkmark & \textbf{3.54} & \textbf{32.73} & \textbf{4.64} & \textbf{73.07} & \textbf{5.11}  \\

    \hline
    \end{tabular}
    \vspace{-9pt}

\end{table*}

We select the public datasets EYEDIAP \cite{funes2014eyediap} and MPIIFaceGaze \cite{zhang2017mpiigaze}  to compare the performance of our method with the existing ones. 
We additionally build a new dataset named IMRGaze to validate the effectiveness of our model with the Easy-Calibration module. The details of datasets are as follows.

\textbf{EYEDIAP.}
This dataset consists of continuous video recordings from 16 subjects, which is categorized into 2 subsets based on the gaze targets: 3D floating targets (denoted as EYEDIAP-Float) and 2D screen targets (denoted as EYEDIAP-Screen).
We sample one image 
per 10 frames from the videos, resulting in $\sim$30K images (10K from EYEDIAP-Float and 20K from EYEDIAP-Screen) with $640 \times 480$ resolution. To enable the conversion between PoG and PoGz, we derive the CCS-to-SCS transformation parameters according to the given 3D coordinates of gaze targets in the EYEDIAP-Screen subset.

\textbf{MPIIFaceGaze.}
This dataset 
contains data from 15 subjects in natural environments, with an average of 2,500 images per subject. The images are collected by laptops in daily life, and the CCS-to-SCS transformation parameters are provided, which is required to derive the PoG. 

\textbf{IMRGaze.}
Our IMRGaze dataset consists of video data collected from 12 subjects. The dataset is divided into two parts: General data for training and evaluating the model; Calibration data for validating the effectiveness of the Easy-Calibration module on improving model performance.

General data: The collection process follows EYEDIAP's floating target scenario. Subjects gaze at a floating ball while maintaining static head poses or free head movements. An RGB-D camera (Intel® RealSense™ LiDAR Camera L515) captures the RGB and depth videos at 30 fps, with resolution of $640 \times 480$ for RGB and $320 \times 240$ for depth, which provides real-time 3D information of both facial features and target positions for head pose and gaze vector computation.
The videos undergo identical processing as EYEDIAP, with MediaPipe \cite{48292} for facial landmark detection and manual verification by experienced annotators. This yields $\sim$10K normalized images, as shown in Fig. \ref{fig:imr}(a). The distribution of gaze direction and head pose is presented in Table \ref{tab:distribution}.

Calibration data: The collection process of calibration data is illustrated in Fig. \ref{fig:calibration}, which contains rich facial and gaze-related information.
Labels are generated according to Eq. ~\ref{eq:cali-gt}. The calibration videos undergo identical preprocessing as EYEDIAP, resulting in an average of 100 calibration images per subject, as demonstrated in Fig. \ref{fig:imr}(b).
All the subjects have signed an informed consent about the experiments before they participate in. This study has been approved by the Institutional Review Board for Human Research Protections of Shanghai Jiao Tong University (No. B20240244I).

\subsection{Evaluation Metrics}
For the gaze direction $\bm{g}_o$ and $\bm{g}_n$, we adopt the angle error as the evaluation metric. Let $\bm{g}$ denote the ground truth, and $\hat{\bm{g}}$ denote the prediction. The angular error can be calculated as: 
\begin{equation}
E(\bm{g},\hat{\bm{g}}) = \arccos\frac{\bm{g} \cdot \hat{\bm{g}}}{\|\bm{g}\| \|\hat{\bm{g}}\|}
\end{equation}\label{eq:error}

For the PoG, we use the Euclidean distance between the ground truth $O$ and the prediction $\hat{O}$ on the target plane as the evaluation metric.
\begin{equation}
 E(O,\hat{O}) = \sqrt{(x- \hat{x})^2 + (y- \hat{y})^2}  
\end{equation}
where the $(x,y)$ is the 2D coordinates of the PoG. 

\subsection{Training Configuration}

For EYEDIAP, we set the loss weights as $\lambda_{\bm{g}_n}\!=\!1$, $\lambda_{\bm{g}_o}\!=\!\lambda_{\bm{r}_o^n}\!=\!0.5$, $\lambda_{\mathrm{PoGz}}\!=\!0.1$, $\lambda_{\mathrm{face}}\!=\!0.1$. The model is trained on an NVIDIA 3080 Ti GPU with a batch size of 128. We use the Adam optimizer with an initial learning rate of 0.0001.
For MPIIFaceGaze, the loss weights follow those of EYEDIAP, but the batch size is 256 and the initial learning rate is 0.001. 
We conduct 4-fold cross-validation on EYEDIAP and leave-one-out on MPIIFaceGaze, following most studies \cite{cheng2024appearance}.

For IMRGaze, the training and validation configurations are consistent with those used on EYEDIAP. As for the calibration fine-tuning: First, we obtain the best validation model for each fold and save its parameters. Subsequently, we use 50\%  and 100\% of the calibration data (approximately 50 and 100 images per subject on average) for personal calibration. We load the saved model parameters and fine-tune the model using the calibration data. The fine-tuning is conducted with an initial learning rate of 0.00001. Finally, the fine-tuned model is evaluated on the test set corresponding to that fold.

\subsection{Ablation Study}
To examine the contribution of different branches in the multi-task architecture, we perform ablation studies by removing specific branches, including the $\bm{g}_o$ and PoGz branches illustrated in Fig. \ref{fig:the whole pipeline} by setting the weights $\lambda_{\bm{g}_o}\!=\!0$ or $\lambda_{\mathrm{PoGz}}\!=\!0$ and freezing the corresponding decoders. We then compare the angular error of $\bm{g}_n$ in these ablated models with our basic model (without Easy-Calibration) and the full model on EYEDIAP-Float and IMRGaze datasets.

\section{Results and Analysis}


\subsection{Comparison on Public Datasets}


\begin{figure*}[t]
    \centering
    \includegraphics[width=0.99\linewidth]{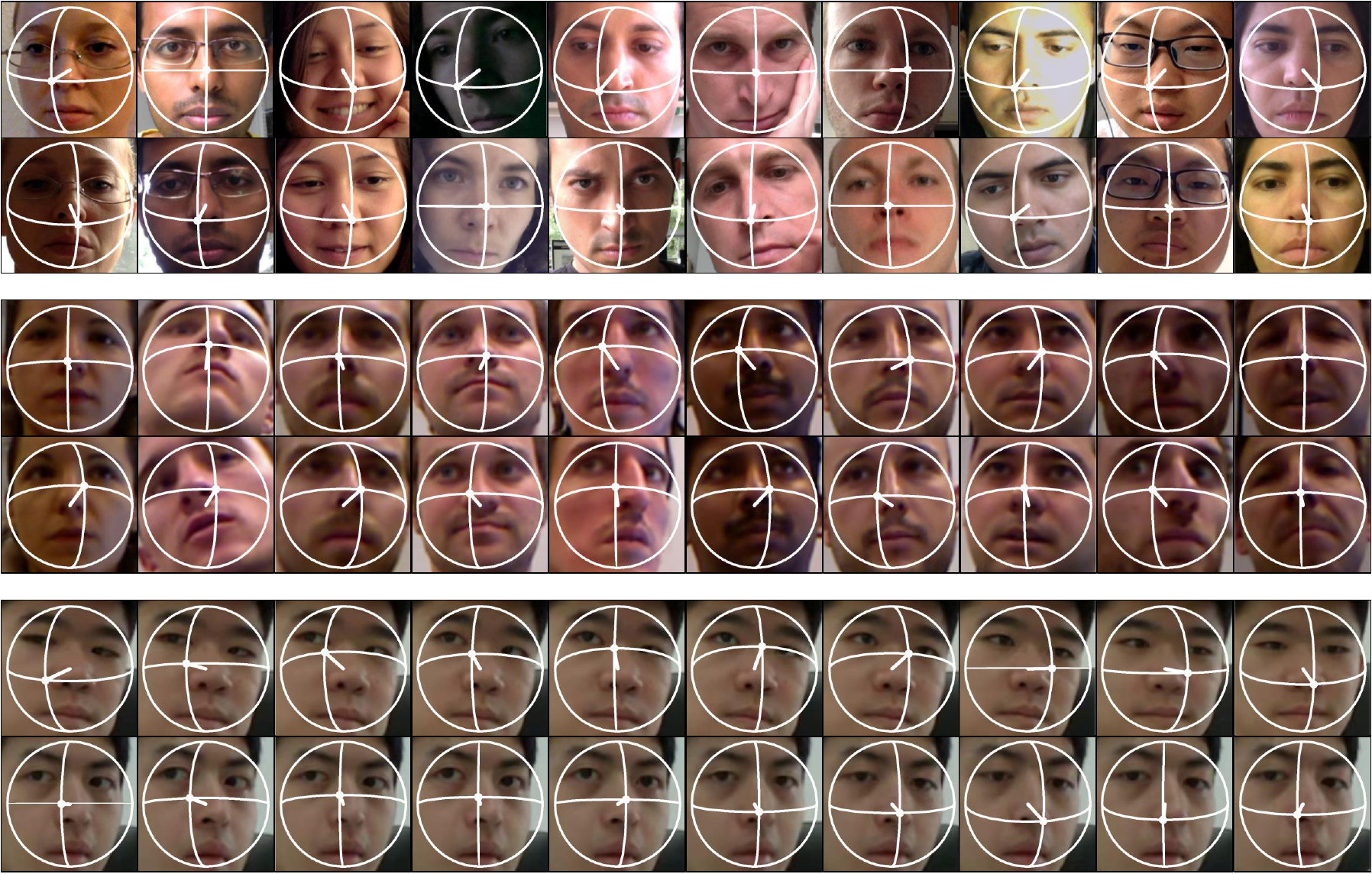}
    \vspace{-3pt}
    \caption{Qualitative results. The row 1-2, 3-4, 5-6 show the subjects and the predicted gaze directions from the MPIIFaceGaze, EYEDIAP and IMRGaze datasets, respectively. In each image, the arrow represents the gaze direction, and the sphere helps to better visualize the 3D effects.}
    \label{fig:qualitative result} 
    \vspace{-9pt}
\end{figure*}

We compare the performance of our basic model on the public datasets MPIIFaceGaze, EYEDIAP-Screen, and EYEDIAP-Float with the existing gaze estimation methods. We select GazeTR \cite{BM:cheng2022gaze}, AGE-Net \cite{BM:shi2024agent}, RCNN \cite{BM:palmero2018recurrent}, L2CS-Net  \cite{abdelrahman2023l2cs} that are proposed for the gaze direction prediction task, and iTracker \cite{Krafka_2016_CVPR}, AFF-Net \cite{BM:bao2021adaptive} that are proposed for the PoG prediction task.
The results are presented in Table \ref{tab:BMscreen}. Since the lack of calibration data in these datasets, we can only present the results of our basic model without the Easy-Calibration module. For the methods \cite{Krafka_2016_CVPR,BM:bao2021adaptive}, we report the experimental results from the review paper \cite{cheng2024appearance}. 

For the prediction of gaze direction, our proposed MAGE outperforms other existing methods across all the three datasets. Specifically, on the MPIIFaceGaze dataset, MAGE achieves the lowest angular error of 3.54\textdegree, superior to the second-best method, AGE-Net, by 7.33\%. On the EYEDIAP-Screen and EYEDIAP-Float datasets, MAGE also outperforms all the other methods with only 4.64\textdegree \ and 5.11\textdegree \ angular errors respectively, which suggests the effectiveness and robustness of our proposed model.

For the prediction of PoG, our MAGE also shows promising results. On the MPIIFaceGaze dataset, it achieves an error of 32.73 mm, which is significantly better than the second-best method, AFF-Net, with a PoG error of 42.10 mm. Similarly, on the EYEDIAP-Screen dataset, MAGE achieves a PoG error of 73.07 mm, outperforming other models, including AFF-Net (92.5 mm).

Based on the results presented above, our method not only provides complete gaze information for practical applications, but also outperforms specialized methods designed for predicting gaze direction or PoG, particularly for the latter one. This improvement can be attributed to our network architecture and the multi-task constraints, which effectively exploit the relationship between gaze direction and PoGz, thereby enhancing the overall performance.

\subsection{Comparison Results on IMRGaze Dataset}
\label{sec:BM-IMR}

We report the angular error of normalized gaze $\bm{g}_n$ as the evaluation metric on IMRGaze, and assess our model MAGE with 50\% and 100\% calibration data, respectively. For comparison, we also evaluate GazeTR, AGE-Net, RCNN and L2CS-Net on the general data of IMRGaze. Results in Table \ref{tab:IMR} show that MAGE outperforms other methods, consistent with its performance on the public datasets.

Our proposed Easy-Calibration module reduces gaze direction error by 0.23\textdegree \ (4.27\%). Notably, MAGE-50\% and MAGE-100\% perform similarly, indicating that 50\% calibration data (approximately 50 images per subject) is sufficient for fine-tuning. For video data, sampling one frame per 10 frames allows the calibration to be completed in roughly 17 seconds, highlighting the efficiency of Easy-Calibration.

\begin{table}[t]
    \vspace{+5pt}
    \centering
    \caption{Comparison of gaze direction estimation on IMRGaze dataset}
    \label{tab:IMR}
    \renewcommand\arraystretch{1.2}
    \begin{tabular}{llc}
    \hline\hline
        Model                                 & Input & Gaze Dir. (\textdegree) \\ \hline

        GazeTR \cite{BM:cheng2022gaze}        & Face & 5.69 \\ 
        AGE-Net \cite{BM:shi2024agent}        & Eyes & 8.29  \\
        RCNN \cite{BM:palmero2018recurrent}   & Face \& Eyes  & 6.71  \\ 
        L2CS-Net \cite{abdelrahman2023l2cs}      & Face & 6.22  \\ \hline
        MAGE-50\% (ours)                    & Face & 5.18  \\
        
        \textbf{MAGE-100\% (ours)}                   & Face & \textbf{5.15}  \\

    \hline
    \end{tabular}

\end{table}

\begin{table}[t]
    \vspace{+6pt}
    \centering
    \caption{Ablated results of our proposed MAGE}
    \renewcommand\arraystretch{1.2}
    \begin{tabular}{lcc}
    \hline\hline
        Model                   & EYEDIAP-Float (\textdegree) & IMRGaze (\textdegree) \\ \hline
        w/o PoGz branch         & 5.29 & 5.83 \\
        w/o $\bm{g}_o$ branch        & 5.19 & 5.58 \\ 
        w/o Easy-Calibration    & 5.11 & 5.38 \\ 
        MAGE (full)           & N/A  & 5.15 \\ 

    \hline
    \end{tabular}
    \label{tab:Ablation}
    \vspace{+1pt}
\end{table}


\subsection{Results of Ablation Study} 
The results of the ablation study are presented in Table \ref{tab:Ablation}. The full MAGE demonstrates superior performance over its ablated variants on IMRGaze, achieving angular error reductions of 0.68\textdegree \ (11.66\%) , 0.43\textdegree \ (7.70\%) compared to the models without the PoGz branch and $\bm{g}_o$ branch, respectively. Similarly, on EYEDIAP-Float, the basic model exhibits improvements of 0.18\textdegree \ and 0.08\textdegree \ in angular error compared to the corresponding ablated versions. This highlights that the prediction of PoGz serves as a crucial source of information for gaze direction estimation, which indicates the rationality of our multi-task architecture design.

\subsection{Qualitative Results}
Fig. \ref{fig:qualitative result} visualizes the predicted gaze directions of our proposed MAGE for several subjects in three datasets. The row 1-2 show the results on the MPIIFaceGaze dataset, which is collected in natural environments with significant variations in lighting, background, and posture of subjects. The row 3-4 and 5-6 display the results from EYEDIAP and IMRGaze. These two datasets have relatively lower image quality as well as large variations in the head pose and gaze direction. 
In particular, we select a continuous series of images from two subjects in IMRGaze for demonstration. Our MAGE achieves good performance across these datasets, highlighting its accuracy, generalization ability and robustness on gaze estimation tasks.

\vspace{+2pt}
\section{Conclusions}
\vspace{+1pt}
In this paper, we propose MAGE, a multi-task architecture that extracts both the directional and positional gaze information from the RGB facial image, making it directly applicable to the real-word HRI scenarios. We introduce a simplified normalization approach named Easy-Norm to preprocess the input data. Our model mainly predicts 2 components: gaze direction and PoGz (gaze point on the $XY$-plane of CCS), which are combined to generate the complete 6-DoF gaze output. 
Furthermore, our Easy-Calibration module is efficient to implement without the need of a screen, 
which utilizes the calibration data to fine-tune our basic model and mitigate the problem of individual variations.
Experimental results demonstrate the SOTA performance of our method on both the gaze direction prediction and the PoG prediction tasks. In the future, we will leverage the transformation relationships among the multi-task outputs to enhance the robustness of the model.
\vspace{-2pt}
\bibliographystyle{IEEEtran}
\bibliography{Ref}

\end{document}